\documentclass[12pt,letterpaper]{article}
\usepackage[margin=1.0in]{geometry}
\usepackage{setspace}

\usepackage{epsfig, amsthm, amsmath, amssymb, latexsym, xpatch, xcolor, mathtools, verbatim, cancel}
\usepackage[titletoc]{appendix}
\usepackage[normalem]{ulem} 
\usepackage{tabularx}
\usepackage{stmaryrd} 
\usepackage{float} 
\usepackage[margin=1cm, font={small,it}]{caption} 

\newcommand{\Cure}{\mathit{Cure}}
\newcommand{\Take}{\mathit{Take}}
\newcommand{\Complier}{\mathit{Complier}}
\newcommand{\Defier}{\mathit{Defier}}
\newcommand{\Always}{\mathit{Always\textit{-}Taker}}
\newcommand{\Never}{\mathit{Never\textit{-}Taker}}
\newcommand{\Individual}{\mathit{U}}

\newcommand{\Assign}{\mathit{Assign}}

\newcommand{\If}{\mathit{if}\,}

\newcommand{\op}{\begin{itemize}}
\newcommand{\ed}{\end{itemize}}

\newcommand{\opp}{\begin{quote}}
\newcommand{\edd}{\end{quote}}

\newcommand{\ope}{\begin{enumerate}}
\newcommand{\ede}{\end{enumerate}}

\newcommand{\im}{\item}

\newcommand{\PP}{\Pr\!}

\newcommand{\given}{\,|\,}

\theoremstyle{plain}

\newtheorem*{lemma*}{Lemma}

\title{The Logic of Counterfactuals and\\ the Epistemology of Causal Inference}

\author{Hanti Lin \\[0.5em] University of California, Davis \\ika@ucdavis.edu}

\begin{document}

\maketitle

\begin{abstract} \noindent 
	The 2021 Nobel Prize in Economics recognized an epistemology of causal inference based on the Rubin causal model (Rubin 1974), which merits broader attention in philosophy. This model, in fact, presupposes a logical principle of counterfactuals, {\em Conditional Excluded Middle} (CEM), the locus of a pivotal debate between Stalnaker (1968) and Lewis (1973) on the semantics of counterfactuals. Proponents of CEM should recognize that this connection points to a new argument for CEM---a Quine-Putnam indispensability argument grounded in the Nobel-winning applications of the Rubin model in health and social sciences. To advance the dialectic, I challenge this argument with an updated Rubin causal model that retains its successes while dispensing with CEM. This novel approach combines the strengths of the Rubin causal model and a causal model familiar in philosophy, the causal Bayes net. The takeaway: deductive logic and inductive inference, often studied in isolation, are deeply interconnected.
 \end{abstract}

\newpage
\section{Introduction}

The debates on inductive inference often proceed in isolation from the controversies about deductive logic, which makes sense. When discussing induction, we often tentatively assume a simple deductive framework, like classical logic, as induction alone poses enough challenges and disagreements. 
But I would like to highlight the intricate relationship between deduction and induction. Specifically, I will explore the connection between the deductive logic of counterfactuals on the one hand, and a very interesting type of induction on the other hand---causal inference. The spotlight will be on the most influential theory of causal inference in health and social sciences. This theory, recognized by the 2021 Nobel Prize in Economics but underdiscussed in philosophy, has been widely used for tasks such as estimating the efficacy of new drugs and the impact of military service on lifetime earnings.

This Nobel Prize-winning theory of causal inference is based on the {\em Rubin causal model} (Rubin 1974), also known as the {\em potential outcome framework}. It is developed on the assumption of a deductively logical principle of counterfactuals:
    \opp 
	\uline{\textsc{Conditional Excluded Middle (CEM)}}
	
	It is logically necessary that 
	\\-\, either $B$ would be the case if $A$ were the case, 
	\\-\, or $B$ would not the be case if $A$ were the case. 
	\edd 
This logical principle is not entirely uncontroversial in science. In fact:
	\op 
	\im Statistician Dawid (2000) raises concerns about this logical principle, leading him to reject the Rubin causal model and its associated theory of causal inference. 
	\im In contrast, computer scientist Pearl (2000) adopts the opposite stance, suggesting that the success of this theory of causal inference supports the theory itself and, in turn, vindicates its underlying logical principle, CEM. 
	\ed 
This debate, mostly pursued in science for now, warrants attention by philosophers. Indeed, Dawid appears unaware that his objection to CEM closely mirrors Lewis's (1973) critique of Stalnaker's (1968) adoption of CEM---a classic debate in philosophy of language. And Pearl is close to offering an indispensability argument for CEM. I will examine both sides of the debate to illustrate how issues of induction intertwine with those of deduction.

In particular, I will first turn Pearl's (2000) preliminary defense of CEM into a full fledged argument. Here is the idea: CEM was already assumed in the early days of the Rubin causal model (Rubin 1974), which found important applications to causal inference in health and social sciences through the work of Imbens \& Angrist (1994) and Angrist, Imbens, \& Rubin (1996), culminating in the 2021 Nobel Prize in Economics. Notably, even though the assumption of CEM was already challenged in the scientific community more than 20 years ago (Dawid 2000), it has remained central to the Rubin causal model to this day. I will explain in detail why CEM has been here to stay for so long. Thus, CEM appears to be an indispensable part of our best scientific theory of causal inference in health and social sciences. An argument for CEM then emerges: an indispensability argument in the style of Quine (1948) and Putnam (1971), as detailed below (Section \ref{sec-lewis}). 

Next, following the good cop/bad cop approach, I will switch sides and undermine the indispensability argument. A new theory of causal inference will be developed to dispense with CEM while preserving the Nobel-Prize-winning applications of the Rubin causal model. The key, somewhat surprisingly, is to combine two causal modeling frameworks: the Rubin causal model, more familiar to health and social scientists, and the causal Bayes net, more familiar in philosophy (Section \ref{sec-date}).

In the final section, \ref{sec-closing}, the good cop/bad cop dialectic will conclude by connecting it to a broader philosophical context, encompassing such topics as the revisability of deductive logic, intertheory relations, and the role of background assumptions in justifying scientific inference.

Before doing all these, I must first tackle a preliminary task. Given the severe lack of discussion on the Rubin causal model in philosophy, I will have to begin by providing an accessible tutorial on it in Section \ref{sec-tutorial}. This tutorial will be accompanied by a fully rigorous version, presented in Appendix \ref{app-rigorous}.

\section{A Gentle Introduction to the Rubin Causal Model}\label{sec-tutorial}

The Rubin causal model has been extensively applied to study various aspects of our medical and economic lives. Think about it: life itself is not unlike a card game.

\subsection{Introducing the Card Game}

There are cards that determine our fates:
	\opp 
		\begin{center} 
		{\bf Card \#1: What If You Took the Treatment?}	
		\end{center}
	Nature gives every individual a card of this form: the back is printed with `$\If \Take = 1$', and the face is printed with `$\Cure = 1$' or `$\Cure = 0$'. 
	\edd 
The former case means that this person would be cured if they took the treatment, while the latter means that this person would {\em not} be cured if they took the treatment. Thus, this card design already presupposes Conditional Excluded Middle. 

There is only one rule for card flipping: any card given to a person is initially face down and will be flipped to reveal the result exactly when the if-clause actually applies to that person.

Similarly, there is also:
	\opp 
		\begin{center} 
		{\bf Card \#2: What If You Didn't Take the Treatment?}	
		\end{center}
	Nature gives every individual a second card, with the back printed `$\If \Take = 0$', and the face is printed with `$\Cure = 1$' or `$\Cure = 0$'. 
	\edd 

Each person's cards \#1 and \#2 define that person's {\em individual treatment effect} (ITE): the value of binary variable $\Cure$ on card \#1 minus its value on card \#2. There are three possible cases:
	\begin{eqnarray*}
	{\rm ITE} 
	&=&
		\begin{cases}
			\begin{aligned}
			1 \quad & (=\, 1-0) && \mbox{i.e. improvement},
			\\
			0 \quad & (=\, 1-1 \mbox{ or } 0-0) &&  \mbox{i.e. no difference},
			\\
			-1 \quad & (=\, 0-1) && \mbox{i.e. deterioration}.
			\end{aligned}
		\end{cases}
	\end{eqnarray*}
The {\em average treatment effect} (ATE) for a population is defined as the average of the individual treatment effects for all individuals in the population. 

A bit of algebra shows that the ATE is equal to the difference between two proportions: 
	\begin{eqnarray*}
	{\rm ATE} &=&
	\mbox{(i) the proportion of `$\Cure = 1$' cards among all cards of type \#1}
	\\[-0.3em]
	&&
	-\; \mbox{(ii) the proportion of `$\Cure = 1$' cards among all cards of type \#2}.
	\end{eqnarray*}
There is a simple and effective way to estimate term (i): randomly flipping some cards of type \#1 in the population---or equivalently, randomly selecting some people in the population and {\em forcing} them to flip their cards of type \#1. Once the faces of those cards are revealed, register the proportion of the occurrences of `$\Cure = 1$', and use it as an estimate of term (i). Term (ii) can be estimated similarly. This procedure for estimating the ATE is the idea behind {\em randomized controlled trials} (RCTs). The problem, however, is that RCTs are often ethically impermissible.

Fortunately, there is a Nobel-Prize-winning solution, which seeks to estimate, not exactly the ATE, but a closely related causal effect---without forcing anyone to do anything.

\subsection{Switching from the ATE to the LATE}

Let's randomly select individuals from the population and then assign each of them to either the treatment or control group by flipping a coin. Here is the thing: anyone in the treatment group is offered the treatment for free, and they decide whether to take it---there is no forcing anyone to do anything. This creates a new type of card:
	\opp 
		\begin{center} 
		{\bf Card \#3: What If You Were Assigned to the Treatment Group?}	
		\end{center}
	Nature gives every individual a card of this form: the back is printed with `$\If \Assign = 1$' (where $1$ means the treatment group), and the face is printed with `$\Take = 1$' or `$\Take = 0$'.
	\edd 
This determines whether the individual would or would not take the treatment if assigned to the treatment group. Similarly: 
	\opp 
		\begin{center} 
		{\bf Card \#4: What If You Were Assigned to the Control Group?}	
		\end{center}
	Nature gives every individual a card of this form: the back is printed with `$\If \Assign = 0$' (where $0$ means the control group), and the face is printed with `$\Take = 1$' or `$\Take = 0$'.
	\edd 
With the new cards, we can define some subpopulations:
	\ope  
	\im {\em Compilers}: those who would take the treatment if assigned to the treatment group, and would not if assigned to the control group (namely, those whose card \#3 and card \#4 are printed with `$\Take = 1$' and `$\Take = 0$', respectively).
	
	\im {\em Defiers}: those who would do the opposite of what compliers would do.
	
	\im {\em Always-Takers}: those who would take the treatment regardless of assignment.
	
	\im {\em Never-Takers}: those who would not take the treatment regardless of assignment.
	\ede  
By Conditional Excluded Middle, those four subpopulations jointly exhaust the entire population. 

Now, let the target of estimation be, not exactly the ATE, but a closely related quantity, the LATE, short for {\em local average treatment effect}. The LATE is defined as the average of the individual treatment effects of just the {\em compliers} in the population, or more formally: 
	\begin{eqnarray*}
	\textrm{LATE} 
	&=_\text{df}& \frac{\sum_{\, i: \,\text{being a complier}} \,\text{ITE of individual } i}{ \text{the number of compliers} }
	\,.
	\end{eqnarray*}	
Interestingly, when there are no defiers and other conditions are met, it is possible to estimate the LATE without forcing anyone to take the treatment, as will be shown shortly.

\subsection{Estimating the LATE}

The standard procedure for estimating the LATE is known as {\em instrumental variable estimation}. To understand it, we need a theorem, now a classic result in econometrics and statistics (Imbens \& Angrist 1994, Angrist, Imbens, \& Rubin 1996):
	\opp 
	{\bf Informal Statement of Theorem 1 (Identification of the LATE).}
	\emph{
	In the card game presented above, which already builds in Conditional Excluded Middle, suppose that the following four assumptions hold:
	\op 
	\im {\sc (Random Selection)} Everyone has an equal probability of being selected.
	\im {\sc (Random Assignment)} The selected people are randomly assigned, with equal probabilities, to the treatment or control group.
	\im {\sc (Existence of Compliers)} There are compliers.
	\im {\sc (No Defiers)} There are no defiers.
	\ed  
	Then the {\rm LATE} can be expressed solely in terms of probabilities over the three observable variables---$\Assign$, $\Take$, and $\Cure$---without counterfactuals. Specifically:
	\begin{eqnarray*}
	{\rm LATE}
	&=& \frac{
	\PP\left(\Cure = 1\given \Assign = 1\right) \,-\, \PP\left(\Cure = 1 \given \Assign = 0\right)
	}{
	\PP\left(\Take = 1 \given \Assign = 1\right) \,-\, \PP\left(\Take = 1 \given \Assign = 0\right)
	} \,.
	\end{eqnarray*}
	}
	\edd 
To be more precise, this equation holds under the assumptions 1-8 formalized in Appendix \ref{app-rigorous}. The first four of those eight assumptions, including CEM, are build into the card design; the remaining four are stated informally as the bullet points in the above.

Some explanations are in order. First, the assumption that there are compliers plays a straightforward role by ensuring that the target of estimation, the LATE, is well-defined (i.e., has a nonzero denominator).

The assumption of no defiers plays a more interesting role: to delineate the scope of application. For example, when estimating the causal effect of a newly designed drug not yet available on the market, no one in the control group could take the new drug, which implies that no one is a defier. Another example comes from Angrist's (1990) now-classic study on the Vietnam War, where ``random assignment'' refers to the draft lottery, ``treatment'' to military service, and the ``medical result'' to lifetime earnings. A defier in this scenario is someone being this crazy: one who would volunteer for military service if they were not drafted but would avoid service if drafted. Here, it is also reasonable to assume that no defiers exist. However, in cases where it is implausible to assume that, the present theorem provides no guidance on estimating causal effects.

Let's now turn to $\PP$, the probability function in use. The probabilities discussed in this paper are restricted to physical objective probabilities. These probabilities might be frequencies (Neyman 1955), propensities (Popper 1959), or primitive physical states posited in science (Sober 2000: sec. 3.2)---to mention just the options developed with classical statistics in mind, which often serves as the background theory for the Rubin causal model. I remain open to the metaphysics of physical objective probabilities; the focus of this paper is epistemology.

The first conditional probability in the equation, $\Pr(\Cure = 1 \given \Assign = 1)$, is defined in the standard way: 
	\begin{eqnarray*}
	\Pr(\Cure = 1 \given \Assign = 1) &=&
	\frac{\Pr(\Cure = 1 \wedge \Assign = 1)}{\Pr(\Assign = 1)}	 \,,
	\end{eqnarray*}
where the denominator is the probability that a randomly selected person is assigned to the treatment group $(\Assign = 1)$, and the numerator is the probability that a randomly selected person is assigned to the treatment group $(\Assign = 1)$  and then gets cured $(\Cure = 1)$. This unknown conditional probability can be easily estimated---by the observed proportion of the cured individuals in the treatment group. The other three conditional probabilities can be similarly estimated by observed proportions. This procedure for estimating the conditional probabilities on the right-hand side of the equation, and thus estimating the LATE on the left-hand side, is known as {\em instrumental variable estimation} (with the variable $\Assign$ serving as the so-called instrument).

This result marks an important achievement. Recall that the LATE is defined in counterfactual terms, using the contents of cards that cannot all be flipped to reveal their faces at the same time---a single person cannot simultaneously take the treatment and not take it.
Fortunately, to estimate the LATE, it suffices to observe some proportions in the treatment and control groups and estimate the counterfactual-free, conditional probabilities on the right-hand side of the equation in Theorem 1. It is amazing that an interesting quantity defined in counterfactual terms (the LATE on the left) can be {\em identified} with a quantity that depends solely on counterfactual-free probabilities (on the right), which are easy to estimate. Thus, this theorem is also known as an {\em identification} result. Many important theorems in statistics and econometrics for causal inference are identification results.


For a rigorous statement of Theorem 1, see Appendix \ref{app-rigorous}, which seeks to improve upon standard presentations. To be sure, there is a particularly lucid and frequently cited  presentation in the statistics article by Angrist, Imbens, \& Rubin (1996, Proposition 1), but those authors list only four assumptions, omitting an explicit statement of CEM. In Appendix \ref{app-rigorous}, I identify eight assumptions in total, including CEM, of course.

This concludes the first task of this paper: a crash course on the Rubin causal model and the identification result for the LATE.


\section{Playing Good Cop}\label{sec-lewis}

The preceding discussion can inspire a new argument for Conditional Excluded Middle. Let me flesh it out, playing the role of the good cop---for now.

\subsection{A New Argument for CEM}\label{sec-warmup}

Why might it be interesting to have a new argument supporting CEM? The reason is that there is a highly influential argument against CEM (Lewis 1973). Let me briefly review it. Consider the following pair of sentences:
	\op
	\im[$(A)$] If $i$ took the treatment, $i$ would be cured.
	\\[-2.3em]
	\im[$(B)$] If $i$ took the treatment, $i$ would not be cured.
	\ed 
CEM requires that the disjunction $(A) \vee(B)$ be true in every possible world. To find a counterexample, consider an indeterministic world in which the following holds:
	\op 
	\im[$(C)$]
	If $i$ took the treatment, $i$ would have a nontrivial probability $p$ of being cured and a probability $1-p$ of being not cured, where nontriviality means that $p$ lies strictly between $0$ and $1$.
	\ed 
Then argue as follows that the truth of $(C)$ implies the falsity of both $(A)$ and $(B)$:
	\begin{center}
		\uline{\textsc{Indeterminist Argument Against CEM}}
	\end{center}
	\op 
	\im[1.] Assume that $(C)$ is true. 
	
	\im[2.] By 1, if the individual $i$ took the treatment, $i$ would have a more-than-zero probability of being not cured.
	
	\im[3.] So, if $i$ took the treatment, $i$ could be not cured. (This follows from 2, by the inference from `would have a more-than-zero probability to be' to `could be'.)
		\op 
	\im[4.] Now, suppose for {\em reductio} that $(A)$ is true: if $i$ took the treatment, $i$ would be cured. 
	
	\im[5.] Then, by 3 and 4, we have: if $i$ took the treatment, $i$ would be cured {\em and} could be not cured---absurd. 
		\ed 
	\im[6.] So, by the {\em reductio} argument from 4 to 5, it follows that $(A)$ is false.
	
	\im[7.] By symmetry, $(B)$ is false, too; thus $(A)$ and $(B)$ are both false.
	\ed 
In a nutshell, nontrivial counterfactual probabilitiy refutes CEM---or so Lewis (1973) concludes. H\'{a}jek (manuscript) further argues that such counterexamples to CEM are pervasive in the actual world we live in.

The above is just round one of the debate. The next round features responses from defenders of CEM, such as Stalnaker (1981). This debate has unfolded across philosophy of language (Williams 2010), metaphysics (Emery 2017), and traditional epistemology (Boylan 2024).\footnote{For reviews of this debate, see Loewenstein (2021) and Mandelkern (2022, sec. 17.3.4).} I submit that philosophy of science is also an area where we can explore a new argument for CEM:
	\opp 
	\begin{center}
		\uline{\textsc{Indispensability Argument For CEM}}
	\end{center}
	\edd 
	\opp CEM is assumed in our best theory of causal inference in health and social sciences, whose application to instrumental variable estimation underpinned the 2021 Nobel Prize in Economics. Despite the influential challenge raised by statistician Dawid (2000) more than twenty years ago in the scientific community ---a challenge very similar to Lewis's (1973) worry from nontrivial counterfactual probability---CEM has persisted as a core assumption of this theory to this day. Thus, CEM seems indispensable. Given that we should believe in our best theory of causal inference in health and social sciences, and that CEM is an indispensable part of it, it seems that we have no choice but to believe in CEM---for fear of {\em intellectual dishonesty}, in Putnam's (1971) terms.	
	\edd 

As just mentioned, the indispensability of CEM is already supported by its persistence in the face of the challenge in the scientific community. This indispensability can be further reinforced by examining the role of CEM in the Rubin causal model, to which I turn now.

\subsection{What's the Role of CEM, Exactly?}\label{sec-role}

CEM has been here to stay for a long time due to its crucial role in proving key lemmas that underpin causal inference. To see this clearly, we need to delve into some formal details of the Rubin causal model. This section is more technical and can be skipped on a first reading.

Let $\Take_{i} = 1$ express the proposition that the individual $i$ takes the treatment. Similarly, $\Cure_{i} = 1$ expresses that $i$ is cured, and $\Assign_{i} = 1$ expresses that $i$ is assigned to the treatment group (rather than the control group). To this notation, we can add superscripts to express counterfactuals, such as the following:
	\op 
	\im $\Cure_{i}^{\Take_{i} = 1} = 1$ means that $i$ would be cured if $i$ took the treatment.
	\im $\Take_{i}^{\Assign_{i} = 0} = 0$ means that the individual $i$ would not take the treatment if $i$ were assigned to the control group. 
	\ed 
The Rubin causal model makes some logical assumptions:
	\opp 
	{\bf Assumption (Centering/Consistency).}\footnote{While `Centering' is the standard name for this logical principle in philosophy, the scientific literature uses `Consistency' instead.} The antecedent of a counterfactual is redundant if it happens to be true; in symbols:
	$$\mathit{X}_i = x \;\Rightarrow\; \left(\mathit{Y}_i^{\mathit{X}_i = x} = y \;\Leftrightarrow\; \mathit{Y}_i = y \right) .
	$$
	\edd 
There is another logical assumption, being the focus of this paper:
	\opp  
	{\bf Assumption (Conditional Excluded Middle, or CEM).} If $Y_i$ is a binary variable, so is the counterfactual variable $Y_i^{X_i = x}$. 
	\edd 
Notably, most presentations in the scientific literature only mention in passing that $Y_i^{X_i = x}$ is a binary variable, making this assumption look more innocuous than it actually is. The substance of this assumption can be appreciated only by going from the formalism back to the intended interpretation: To say that $\Cure_i^{\Take_i = 1}$ is binary is to say that either $\Cure_i^{\Take_i = 1} = 1$ or $\Cure_i^{\Take_i = 1} = 0$, which means that either $i$ would be cured under the treatment or $i$ would not be cured under the treatment---an instance of CEM.

The four cards for each individual $i$ correspond to the four counterfactual variables: $\Cure_i^{\Take_i = 1}$, $\Cure_i^{\Take_i = 0}$, $\Take^{\Assign_i = 1}$, and $\Take^{\Assign_i = 0}$, whose values correspond to the faces of the four cards. Thus, the card-based definitions presented above can be formalized as follows. First, the ITE (individual treatment effect) for an individual $i$ is defined by:
	\begin{eqnarray*}
	\textrm{ITE}_i 
	&=_\text{df}& \Cure_i^{\Take_i = 1} - \Cure_i^{\Take_i = 0} \,.
	\end{eqnarray*}
The four subpopulations are defined as follows:
	\begin{eqnarray*}
	\Complier(i) 
		& \Leftrightarrow_\text{df}&
		\Take_i^{\Assign_i = 0} = 0 \,\wedge\, \Take_i^{\Assign_i = 1} = 1 \,;
	\\
	\Defier(i) 
		& \Leftrightarrow_\text{df}&
		\Take_i^{\Assign_i = 0} = 1 \,\wedge\, \Take_i^{\Assign_i = 1} = 0 \,;
	\\
	\Always(i) 
		& \Leftrightarrow_\text{df}&
		\Take_i^{\Assign_i = 0} = 1 \,\wedge\, \Take_i^{\Assign_i = 1} = 1 \,;
	\\
	\Never(i) 
		& \Leftrightarrow_\text{df}&
		\Take_i^{\Assign_i = 0} = 0 \,\wedge\, \Take_i^{\Assign_i = 1} = 0 \,.
	\end{eqnarray*}
The target of estimation is the local average treatment effect for the compliers, which is defined by:
	\begin{eqnarray*}
	\textrm{LATE} 
	&=_\text{df}& \frac{\sum_{i: \, \Complier(i)} \textrm{ITE}_i }{ \#\{ i :  \Complier(i) \} }
	\,,
	\end{eqnarray*}	
where the denominator denotes the size of the complier subpopulation. We can then derive a probabilistic formula to express the LATE:
	\opp 
	{\bf Lemma A.} Under the assumption of Random Selection (that everyone has an equal probability of being selected), we have: 
	\begin{eqnarray*}
	{\rm LATE} 
		&=& \PP\left( \Cure^{\Take = 1} = 1 \given \Complier \right)
	\; - \;
		\PP\left( \Cure^{\Take = 0} = 1 \given \Complier \right) \,.
	\end{eqnarray*}	
	\edd 
This formula is often treated as a definition in textbooks for convenience (Hern\'{a}n \& Robins 2023), but is actually a lemma in the rigorous treatment (Imbens \& Rubin 2015). If we unpack the conditional probabilities on the right-hand side using the standard definition, there will appear a denominator $\PP\left( \Complier \right)$, the probability of selecting a complier from the population, which by Random Selection equals the proportion of compliers in the population. So, in the existing proofs of the theorem for identifying and estimating the LATE, the crux is find a formula that helps estimate the proportion of compliers in the population. It is in this task that CEM is deeply involved. Let me explain. 

The existing proofs start with this lemma: 
	\opp 
	{\bf Lemma B.} Under the assumption of CEM, the four subpopulations just defined---compliers, defiers, always-takers, and never-takers---are mutually exclusive and jointly exhaustive.
	\edd 
This lemma is easy to prove: mutual exclusion follows immediately from the definitions; joint exhaustion follows immediately from the definitions and the assumption of CEM. Thanks to Lemma B, the following four proportions sum to $1$:
	\op 
	\im[(1)] the proportion of compliers in the population;
	\\[-2.2em]
	\im[(2)] the proportion of defiers in the population;
	\\[-2.2em]
	\im[(3)] the proportion of never-takers in the population;
	\\[-2.2em]
	\im[(4)] the proportion of always-takers in the population.
	\ed 
So, to estimate the primary target, (1), it suffices to subtract the estimates of (2)-(4) from $1$. This is the {\em first} role played by Lemma A, and hence, by CEM. Then, since (2) is equal to zero by the assumption of No Defiers, it remains to estimate (3) and (4).

To estimate (3), consider the following three quantities:
	\op
	\im[$(3)$] the proportion of never-takers in the population;
	\\[-2.2em]
	\im[$(3')$] the proportion of never-takers in the control group;
	\\[-2.2em]
	\im[$(3'')$] the proportion of those who end up not taking the treatment in the control group.
	\ed 
Under the assumption of Random Assignment to the treatment or control group, proportion $(3)$ can be estimated by proportion $(3')$, if we can obtain an accurate value for the latter. And we can. The idea is to exploit this lemma:
		\opp  
		{\bf Lemma C.} Under the assumptions of CEM, Centering, and No Defiers, it follows that, within the treatment group, the never-takers are exactly those who end up not taking the treatment. Or in symbols, $\Assign_i = 1$ implies this equivalence:	
		$$\Never(i)  ~\Leftrightarrow~ \Take_i = 0\,.$$
		\edd  	
Thanks to this lemma, proportion $(3')$ is equal to proportion $(3'')$, which can be easily obtained through observartion: simply count the number of individuals not taking the treatment in the treatment group and divide it by the size of the treatment group. To recap: CEM is assumed in Lemma C, which enables us to use proportion $(3'')$ as an accurate value of proportion $(3')$, which, by Random Assignment, can then be used as a good estimate of proportion $(3)$. Very ingenious indeed!

The idea behind the proof of Lemma C is also clever, drawing on Lemma A. This is a {\em second} role played by Lemma A, and hence, by CEM. Let me present the proof in plain language.

{\em Proof of Lemma C.} To prove the ``$\Rightarrow$'' direction, consider any individual $i$ being a never-taker in the treatment group. Then, by the assumption of Centering, $i$ does not take the treatment. Now, to prove the ``$\Leftarrow$'' direction, consider any individual $i$ in the treatment group who ends up not taking the treatment. By Lemma A, which relies on CEM, this person $i$ must be one of the following: an always-taker, never-taker, defier, or complier---notably, this is the only place where CEM is employed in this proof. Of those four possibilities, three can be eliminated. Specifically, we can eliminate the possibility that $i$ is a defier, by the assumption of No Defiers. We can also eliminate the possibility that $i$ is an always-taker or complier; for the always-takers and complier in the treatment group end up taking the treatment by the assumption of Centering, but $i$ does {\em not} take the treatment. Thus, the only remaining possibility is that $i$ is a never-taker, as desired. Q.E.D.

Now that we know how to estimate proportion (3), the same trick can be used to estimate (4), the proportion of always-takers in the population, by counting the actual takers in the control group, and by applying a similar lemma with a similar proof.\footnote
	{
	This lemma, {\bf Lemma C'}, states that, under the assumptions of CEM, Centering, and No Defiers, we have that, within the control group, the always-takers are exactly those who end up taking the treatment; in symbols, $\Assign_i = 0$ implies this equivalence:	$\Always(i) \Leftrightarrow \Take_i = 1$.
	}
And recall that proportion (2) equals zero. Once estimates of proportions (2), (3), and (4) are obtained as explained above, subtracting them from $1$ yields an estimate of the primary target, (1), the proportion of compliers. 

\subsection{Wrapping up the Indispensability Argument}

I hope the above reconstruction illuminates the deeply involved roles that CEM plays in the Rubin model and in its applications to causal inference. No wonder CEM has remained a core assumption for more than twenty years even after the influential challenge posed by statistician Dawid in 2000. This strongly suggests that CEM is indispensable to our best theory of causal inference in health and social sciences.

I have thus completed my second task: presenting a new argument that proponents of CEM can explore and utilize---an indispensability argument drawn from the 2021 Nobel Prize in Economics. To further the dialectic, it is now time for me to switch sides and assist opponents of CEM.

\section{Playing Bad Cop}\label{sec-date}

In my role as the bad cop, I undermine the indispensability argument by showing how the above theory of causal inference can be reformulated without the assumption of CEM. This is akin in spirit to what Field (1980/2016) did to challenge the indispensability argument for mathematical realism; he reformulated Newtonian mechanics without referring to real numbers. 

The assumption of CEM is dxeeply involved in the original card game, as we have seen. So, to remove that assumption, the base game needs an overhaul, to be achieved by two expansion packs.

\subsection{An Expansion Pack: Going Stochastic}

In the base game, everyone is given only a {\em single} card printed with `$\If \Take = 1$', whose face determines whether that person would, or would not, be cured under the treatment. But now imagine that you are given not just one card printed with `$\If \Take = 1$', but a {\em deck} of such cards, where 80\% are printed with `$\Cure = 1$' on their faces, and the remaining 20\% with `$\Cure = 0$'. Let this deck be thoroughly shuffled, with all faces down initially. What if you took the treatment? Nature would then randomly draw a card from this deck and flip it to reveal your medical result. Consequently, you would have an 80\% probability of being cured.\footnote{If randomly drawing a card from a deck does not sound chancy enough, replace it with measuring an observable in a quantum-mechanical system.} So, you could be cured and could be not cured, and hence, it is neither true that you would be cured nor that you would not be cured. CEM is thereby rendered invalid---or so the Lewisians contend. 

Let's generalize. In the base game, every individual is given four cards, answering the following what-if questions: 
	\op 
	\im What if one took (or didn't take) the treatment? 
	\\[-2.2em]
	\im What if one were assigned to the treatment (or control) group? 
	\ed 
Now, let each individual's four cards be replaced by four decks, which provide answers in the following form: `If individual $i$ were \ldots, then $i$ would have a probability $p$ of being \ldots'. Such a $p$ is a counterfactual probability---a probability under a counterfactual condition.

So, we now have a stochastic version of the Rubin causal model: single cards are replaced by decks of cards---that is, deterministic outcomes are replaced by counterfactual probabilities. These counterfactual probabilities can then be used to redefine several concepts in the original Rubin causal model. 

Start with the ITE (individual treatment effect). Each individual $i$ still has an ITE, but it is now redefined as the difference between two counterfactual probabilities, or equivalently, two proportions in decks of cards:
	\begin{eqnarray*}
	{\rm ITE}_{\,i}  
	&=_\text{df}&
	\mbox{(i) the proportion of `$\Cure = 1$' cards in $i$'s deck for `$\If \Take = 1$'}
	\\[-0.2em]
	&&
	-\; \mbox{(ii) the proportion of `$\Cure = 1$' cards in $i$'s deck for `$\If \Take = 0$'}.
	\end{eqnarray*}
In the limiting case where each deck contains only one card, the newly defined ITE reduces to the original ITE. 

Subpopulations are redfined, too. Every individual $i$ now has a {\em degree of compliance} ${\rm DC}_{i}$, defined by how one's counterfactual probability of taking the treatment would change if one switched from the control group to the treatment group: 
	\begin{eqnarray*}
	{\rm DC}_{i}
	&=_\text{df}&
	\mbox{$(a)$ the proportion of `$\Take = 1$' cards in $i$'s deck for `$\If \Assign = 1$'}
	\\[-0.2em]
	&&
	-\; \mbox{$(b)$ the proportion of `$\Take = 1$' cards in $i$'s deck for `$\If \Assign = 0$'}.
	\end{eqnarray*}					
The difference between term $(a)$ and term $(b)$ can be positive, zero, or negative, corresponding to three subpopulations:
	\op 
	\im If ${\rm DC}_i > 0$, one is called a {\em complier} (in the general sense). 
	
	\im If ${\rm DC}_i < 0$, one is called a {\em defier}  (in the general sense).

	\im If ${\rm DC}_i = 0$, one is called an {\em indifferent-taker}, with two special cases: an {\em always-taker}, who has $(a)$ = $(b)$ = 100\%, and a {\em never-taker}, who has $(a)$ = $(b)$ = 0\%. 
	\ed

As to the target of estimation, LATE, it is replaced by a more general concept: a weighted average of the individual treatment effects, where each individual's weight $w_i$ is proportional to their degree of compliance ${\rm DC}_i$. This new concept is called the {\em d}egree-of-compliance-weighted {\em a}verage {\em t}reatment {\em e}ffect, or DATE for short. In symbols:
	\begin{eqnarray*}
	{\rm DATE}
	&=_\text{df}&
	\sum_{i: \text{ being a complier}} w_i \, {\rm ITE}_{\,i} \,,
	\\[0.5em]
	w_i &=_\text{df}& \frac{{\rm DC}_{i}}{ \sum_{j: \text{ being a complier}} {\rm DC}_{j} }  \,,
	\end{eqnarray*}						
where the denominator in the definition of weights $w_i$ is a normalizing factor introduced to ensure that the weights sum to $1$.

The present setting is quite general, encompassing the original card game as a limiting case, where every deck contains only one card. In this special case, all compliers are equally compliant, with a maximal degree of compliance ($100\%$ minus $0\%$), which reduces the DATE to the LATE.

\subsection{The Final Expansion Pack: A Causal Bayes Net}

The next step is to state a key assumption in instrumental variable estimation, which, when expressed in plain language, asserts the following:
	\opp
	{\bf Assumption (Instrumentality, Informal Version).} The assignment mechanism (to the treatment/control group) causally influences the medical outcome only through whether an individual takes the treatment. Moreover, there is no common cause shared by the assignment mechanism and the medical outcome.
	\edd 
When this assumption holds, the variable $\Assign$ is called an {\em instrument}. This informal statement is often found in textbooks (Hern\'{a}n \& Robins 2023, sec. 16.1), but interestingly, the standard formalization of this assumption in the Rubin causal model appears quite different, as you can see from the statement of Assumption 2 in Appendix \ref{app-rigorous} (see also Hern\'{a}n \& Robins 2023, technical point 16.1). 

\begin{figure}[H]
	\centering \includegraphics[width=.55\textwidth]{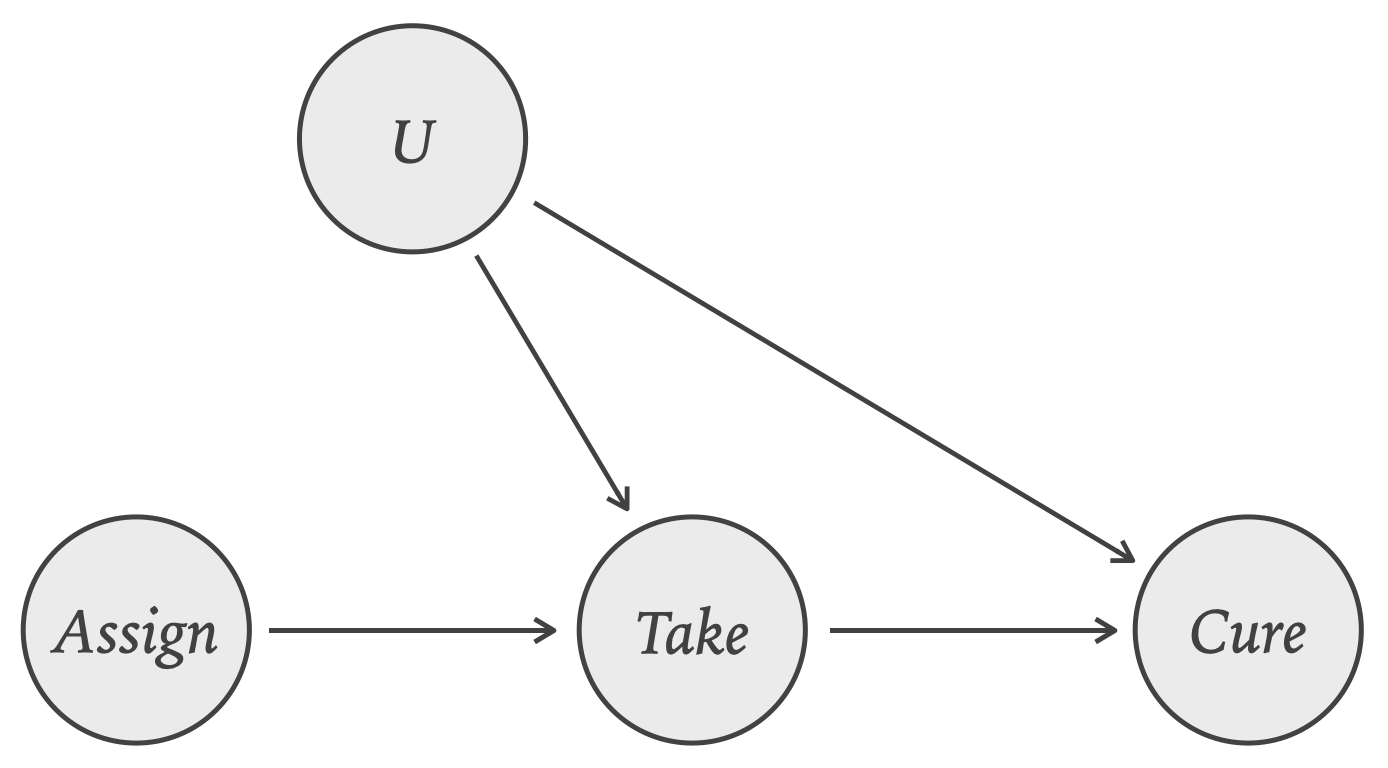}
	\caption{\em The causal structure that captures the Instrumentality assumption}
	\label{fig-dag}
	\end{figure}

I propose a more straightforward formalization of this assumption, using the causal structure depicted in Figure \ref{fig-dag}.
This causal structure is an exact representation of the Instrumentality assumption: every path from the $\Assign$ variable to the $\Cure$ variable passes through the $\Take$ variable, and there is no common cause shared by $\Assign$ and $\Cure$. The confounding variable, $\Individual$, is set to be as fine-grained as possible to avoid missing any confounding factors: its possible values are the individuals in the population. This suffices to encompass all the social, economic, and health conditions of each individual.

\begin{figure}[ht]
	\centering \includegraphics[width=.85\textwidth]{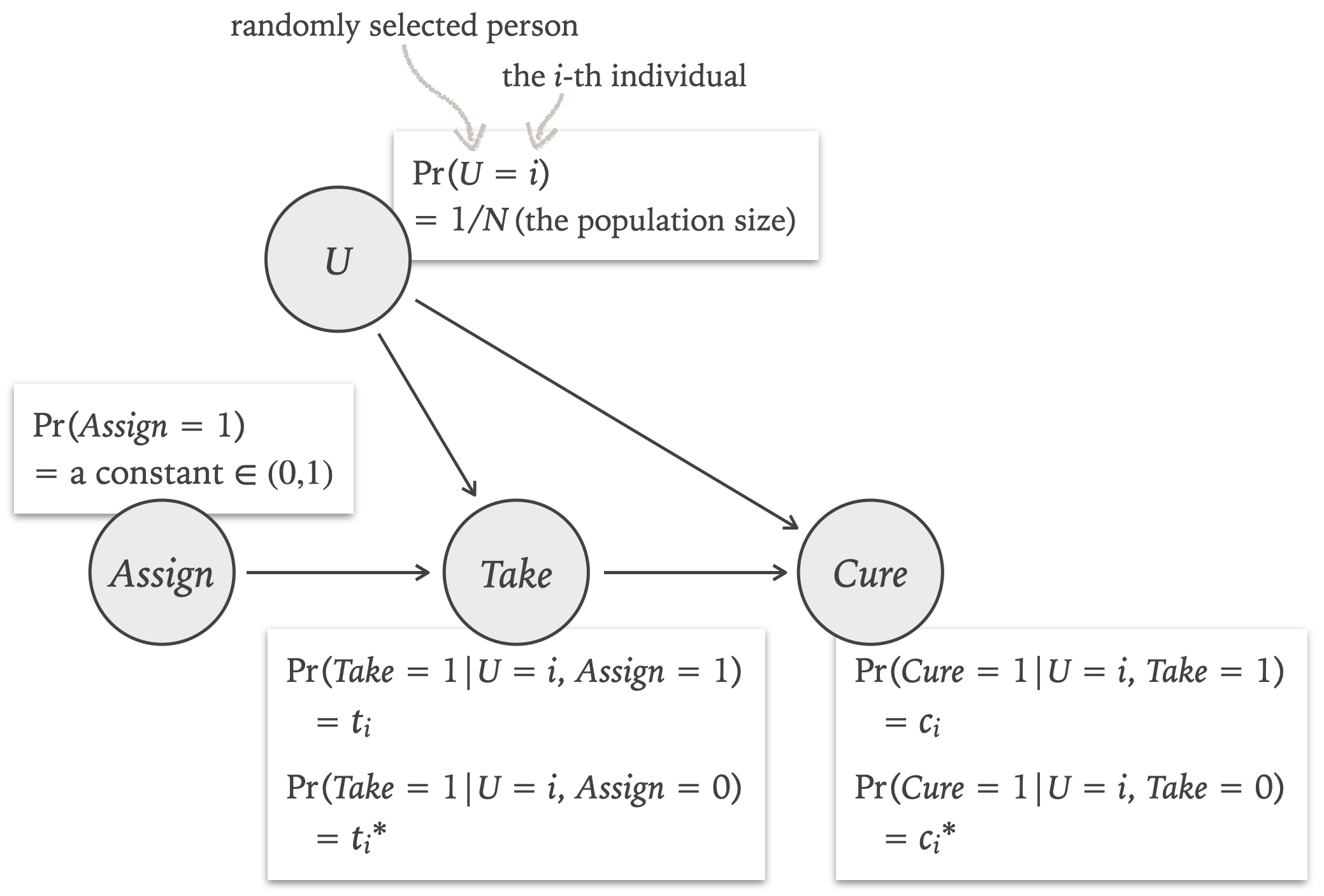}
	\caption{\em The causal Bayes net assumed in Theorem 2}
	\label{fig-cbn}
	\end{figure}

Next, let's turn this causal graph into a {\em causal Bayes net}.\footnote 
	{
	The word `Bayes' can be misleading. Despite the established name in the literature, there is nothing inherently Bayesian in causal Bayes nets, also known as causal Bayesian networks. The probabilities in such networks are most naturally interpreted as physical objective probabilities, measuring the propensities or tendencies of causal influences, rather than degrees of belief.
	}
This is done by specifying some probabilities: the probability distribution of each exogenous variable (i.e., $\Individual$ and $\Assign$), and the conditional probability distribution of each effect variable given its direct cause variables, as shown in Figure \ref{fig-cbn}.

Those probabilities are defined as follows. First, everyone in the population has an equal probability of being selected, so $\PP\,(\Individual = i) = 1/N$, where $i$ is the $i$-th individual and $N$ is the population size. Once a person $i$ is selected, a coin is flipped to decide whether to assign that person to the treatment or control group, with $\PP\,(\Assign = 1) = 1/2$, or more generally, $\PP\,(\Assign = 1)$ being a constant, independent of the individual selected. Finally, the conditional probabilities of effects given direct causes are identified with the appropriate counterfactual probabilities $c_i, c^*_i, t_i$, and $t^*_i$ (as shown in the Figure \ref{fig-cbn}), whose values are taken from the stochastic version of the Rubin causal model, or equivalently, the stochastic expansion pack to the base game:
	\begin{eqnarray*}
	t_i 
		&=_\text{df}&	\mbox{the proportion of `$\Take = 1$' cards in $i$'s deck for for `$\If \Assign = 1$'} \,;
	\\
	t^*_i 
		&=_\text{df}&	\mbox{the proportion of `$\Take = 1$' cards in $i$'s deck for for `$\If \Assign = 0$'} \,;
	\\
	c_i 
		&=_\text{df}&	\mbox{the proportion of `$\Cure = 1$' cards in $i$'s deck for for `$\If \Take = 1$'} \,;
	\\
	c^*_i 
		&=_\text{df}&	\mbox{the proportion of `$\Cure = 1$' cards in $i$'s deck for for `$\If \Take = 0$'} \,.
	\end{eqnarray*}
The main idea can be summarized as follows:
	\opp {\bf Proposal of a New Causal Modeling.} While the original Rubin causal model allows only deterministic outcomes for an individual, it is updated with an expansion pack---replacing single cards with decks---to allow stochastic outcomes with nontrivial counterfactual probabilities. These probabilities are then incorporated into an appropriate causal Bayes net.
	\edd 
This is a combination of two frameworks for causal modeling: the Rubin causal model, more familiar to health and social scientists, and the causal Bayes net, more familiar to philosophers and computer scientists. You will soon see that these two causal models are stronger together, at least for the purpose of pursuing freedom from CEM.

\subsection{Dispensing with CEM}

Finally, we arrive at a new result---a stochastic counterpart to the previous theorem:
	\opp {\bf Theorem 2 (Identification of the DATE).}
	{\em 
	Suppose that the following assumptions hold: 
		\op  
		\im {\sc (Random Selection)} Individuals are randomly selected from the population with equal probabilities.
		
		\im {\sc (Random Assignment)} The selected people are randomly assigned to the treatment or control group with a constant bias strictly between $0$ and $1$ $($e.g., by flipping a fair coin$)$.
		
		\im {\sc (Instrumentality*)} The true causal model is the causal Bayes net depicted in Figure \ref{fig-cbn}.
		
		\im {\sc (Existence of Compliers*)} There are compliers in the population, in the sense that someone's degree of compliance is positive.
		
		\im {\sc (No Defiers*)} There are no defiers in the population, in the sense that no one's degree of compliance is negative. 
		\ed  
		Then the {\rm DATE} can be expressed solely in terms of probabilities over the observable variables---$\Assign$, $\Take$, and $\Cure$---without counterfactuals. Specifically:
	\begin{eqnarray*}
	{\rm DATE}
	&=& \frac{
	\PP\left(\Cure = 1\given \Assign = 1\right) \,-\, \PP\left(\Cure = 1 \given \Assign = 0\right)
	}{
	\PP\left(\Take = 1 \given \Assign = 1\right) \,-\, \PP\left(\Take = 1 \given \Assign = 0\right)
	} \,.
	\end{eqnarray*}
	}
	\edd 

See Appendix \ref{app-proof-2} for a proof. The first two assumptions are actually redundant, as they are already encapsulated in the causal Bayes net posited in the third assumption; but they are stated here to highlight the role of randomization. The last three assumptions are labeled with asterisks to distinguish them from their counterparts in the original Rubin causal model, as stated in Appendix \ref{app-rigorous}.

This new theorem has a notable feature: the right-hand side of the equation for the DATE in the new theorem is {\em identical} to that for the LATE in the classic result. Both are expressed as the same combination of conditional probabilities: $\frac{
	\PP\,\left(\Cure = 1\given \Assign = 1\right) \,-\, \PP\,\left(\Cure = 1 \given \Assign = 0\right)
	}{
	\PP\,\left(\Take = 1 \given \Assign = 1\right) \,-\, \PP\,\left(\Take = 1 \given \Assign = 0\right)
	}$. 
This feature is crucial. Scientists can continue using the same procedure of instrumental variable estimation---estimating the left-hand side by estimating the exact same conditional probabilities on the right-hand side, based on the exact same proportions observed in the treatment and control groups. However, thanks to this new theorem, the old estimation procedure no longer assumes CEM and can be {\em reinterpreted} as estimating the new left-hand side: the newly defined causal effect DATE, of which the LATE is merely a limiting case in a deterministic world (at least for Lewisans). 

This reinterpretation undermines the indispensability argument. Medical and social scientists have practiced instrumental variable estimation for decades, with the stated goal of estimating the LATE under the assumption of CEM. Yet this well-established practice can now be reinterpreted as actually estimating the DATE all along---without assuming CEM. So, the successes of the original theory for causal effect estimation are preserved in the new theory, which dispenses with CEM. The indispensability argument is thus defused.

At this point, proponents of CEM might reply that even if they are compelled to adopt the new theory of causal inference, this would not to stop them from holding onto CEM. Indeed, the assumptions of the new theory only involve counterfactual probabilities and do not explicitly refer to the logic of counterfactuals. And Stalnaker (1981) already argued that one can coherently embrace nontrivial counterfactual probabilities and insist on CEM at the same time. The idea is based on a semantic technique known as {\em supervaluation}, used to resist Lewis's (1973) argument that nontrivial counterfactual probability refutes CEM.

Setting aside the details of supervaluation, it suffices to note that I, as the bad cop at this moment, can concede the points that Stalnakerians made in the previous paragraph. Even so, my main point remains: thanks to the new theory of causal inference, CEM is no longer {\em indispensable}, even if it might still be {\em optional}. This is sufficient to undermine the indispensability argument---the mere optionality of an option is too weak to entail that we should take that option. This concludes my role as the bad cop.

\section{Closing}\label{sec-closing}

I demonstrated how the Rubin causal model could be used to construct a new argument for Conditional Excluded Middle (CEM)---an indispensability argument. Then, I switched sides and undermined that argument. The assumption of CEM is removed by, first, turning the Rubin causal model into a stochastic version and, second, incorporating it into a causal Bayes net. Where does my heart lie---on the Stalnakerian side supporting CEM or the Lewisian side opposing it? You might have guessed that I lean toward the latter, but that is secondary for now. The more important message is this: while the Nobel Prize-winning theory of causal inference has been largely overlooked in philosophy, it actually offers a rich source of interesting issues for philosophers to explore. Let me mention three.


First of all, the dialectic developed above suggests an interesting case for the revisability of logic. If health and social scientists can be persuaded to abandon CEM, possibly following the new theory of causal inference developed above, it would be an example of how empirical inquiry can drive revisions in deductive logic---precisely the kind of case Quine (1951) envisioned. This would underscore the possibility of revising logic in light of not only empirical inquiries but also practical concerns, such as those in the health and social sciences---a much more relatable example than Putnam's (1968) proposal to shift from classical to quantum logic.

So much for deductive logic, but there is also something here for theorists of induction. When scientists justify inductive methods, they rely heavily on their contexts of inquiry, including background assumptions. Past discussions have mostly focused on background assumptions that are physical (Longino 1979, Christensen 1997), methodological, or ethical (Reiss 2020), rather than logical. But do scientists have to assume a logical principle like CEM to justify certain causal inferences? As we have seen, the search for an answer is far from trivial. Thus, background assumptions about deductive logic warrant greater attention from theorists of induction.

Last but not the least, there is also something for those more interested in scientific modeling rather than inference, whether deductive or inductive. Consider the interplay between three approaches to causal modeling:
	\op
	\im[(1)] Rubin causal models (Rubin 1974),
	\\[-2.2em]
	\im[(2)] structural equation models (Pearl 2009),
	\\[-2.2em]
	\im[(3)] causal Bayes nets (Spirtes et al. 2000).
	\ed 
Pearl (2009) famously argues that the first two approaches---Rubin causal models and structural equation models---are essentially equivalent, with the common commitment to deterministic generation of outcomes. I am happy to grant him this point. Yet Pearl further argues that the first two (equivalent) approaches can be used to do everything we can do with the third approach---causal Bayes nets---and this is where I disagree with Pearl. The new theorem suggests that, in at least one important application (instrumental variable estimation), causal Bayes nets generalize Rubin causal models while dropping the metaphysical assumption of deterministic outcomes and removing the logical assumption of Conditional Excluded Middle. This prompts a reconsideration of some questions: Which approach to causal modeling is more general? Which are equivalent, and in what sense? These questions would make for an interesting case study on an important topic: intertheory relations, a subject whose case studies have thus far been largely drawn from natural sciences.\footnote
	{
	For a review of this subject, see Palacios (2024).
	}
I submit that more attention be directed to the relations among causal models in computer science and health and social sciences. While initial steps have been taken by Markus (2021) and Weinberger (2023), their work does not consider causal Bayes nets. Much more remains to be explored.

The Rubin causal model has been overlooked in philosophy for far too long. I hope to have demonstrated that it offers a rich and promising landscape for exploration. It may be surprising that a core issue in philosophy of language (regarding CEM) is deeply connected to philosophy of health and social sciences.


\section*{Acknowledgements} I am indebted to the participants of the Workshop on the Philosophy, Psychology, and Computer Science of Causation held in Kyoto (June 24-26, 2023, Kyoto, Japan), the Conference on Causality in Epidemiology (May 2-4, 2024, Linz, Austria), and the Causation session at the 2024 Philosophy of Science Association Biennial Meeting (November 14-17, 2024, New Orleans, LA, USA). I am especially grateful to Christopher Hitchcock, Peng Ding, Jiji Zhang, Frederick Eberhardt, Jun Otsuka, Xiao-Li Meng, Konstantin Genin, Conor Mayo-Wilson, Tom Wysocki, and Jennifer Jhun for stimulating questions and discussions. 

\appendix

\section{Appendices}

\subsection{The Formalism of the Rubin Causal Model}\label{app-rigorous}

The Rubin causal model builds on a simple idea: ordinary variables are extended to variables under counterfactual conditions, also known as {\em potential outcomes}. Recall that $\Take_{i} = 1$ expresses the proposition that individual $i$ takes the treatment. Similarly, $\Cure_{i} = 1$ says that $i$ gets cured, and $\Assign_{i} = 1$ says that $i$ is assigned to the treatment group (rather than the control group). Given a variable $X_i$, we can use $X_i^{\rm C}$ to denote a {\em potential outcome}, which represents the value of $X$ that individual $i$ would have under the counterfactual condition ${\rm C}$. Here is a substantive assumption:
	\opp 
	{\bf Assumption 1 (Stable Unit Treatment Value, or SUTVA).} The values of the variables of each individual (or unit) are determined independently of the values of the variables of any other individuals. That is, for any variable $X_i$ and any conditions ${\rm C}_1, \ldots, {\rm C}_n$ concerning individuals from $1$ to $n$, we have $X_i^{{\rm C}_1, \ldots, {\rm C}_n} = X_i^{{\rm C}_i}$.
	\edd 
So, this assumption helps simplify the antecedents of counterfactuals. It might be violated in some cases, such as when dealing with a contagious disease in a densely populated community. A further simplification is enabled by the next assumption:
	\opp
	{\bf Assumption 2 (Instrumentality).} For each individual $i$, $\Assign_i$ is an instrumental variable in the following sense: the value of $\Cure_i$ is determined once the value of $\Take_i$ is determined, independently of the value of $\Assign_i$. That is, $\Cure_i^{\Assign_i = a, \Take_i = t} = \Cure_i^{\Take_i = t}$, which omits the assignment $\Assign_i = a$ in the counterfactual condition.
	\edd 
Thanks to the above two assumptions, now we only need to consider just four potential outcomes for each individual $i$: $\Cure_i^{\Take_i = 1}$, $\Cure_i^{\Take_i = 0}$, $\Take_i^{\Assign_i = 1}$, $\Take_i^{\Assign_i = 0}$, which correspond to the four cards that $i$ has in the game presented in the tutorial (Section \ref{sec-tutorial}). 

The design of four cards (as opposed to the four-deck design in my expansion pack) also comes with {\em logical} assumptions:
	\opp 
	{\bf Assumption 3 (Centering/Consistency).} An antecedent, if true, is always redundant; that is, it must be that
	$\mathit{X}_i = x \;\Rightarrow\; \left(\mathit{Y}_i^{\mathit{X}_i = x} = y \;\Leftrightarrow\; \mathit{Y}_i = y \right).
	$
	
	{\bf Assumption 4 (Conditional Excluded Middle, or CEM).} If $Y_i$ is a binary variable, then the counterfactual variable $Y_i^{X_i = x}$ is always a binary variable, too; in other words:
	$$
	Y_i^{X_i = x} = 1 ~\vee~ Y_i^{X_i = x} = 0 \,.
	$$ 
	\edd 
Under the assumption of CEM, the four subpopulations defined below are mutually exclusive and jointly exhaustive (as stated in Lemma A in Section \ref{sec-role}):
	\begin{eqnarray*}
	\Complier(i) 
		& \Leftrightarrow_\text{df}&
		\Take_i^{\Assign_i = 0} = 0 \,\wedge\, \Take_i^{\Assign_i = 1} = 1 \,;
	\\
	\Defier(i) 
		& \Leftrightarrow_\text{df}&
		\Take_i^{\Assign_i = 0} = 1 \,\wedge\, \Take_i^{\Assign_i = 1} = 0 \,;
	\\
	\Always(i) 
		& \Leftrightarrow_\text{df}&
		\Take_i^{\Assign_i = 0} = 1 \,\wedge\, \Take_i^{\Assign_i = 1} = 1 \,;
	\\
	\Never(i) 
		& \Leftrightarrow_\text{df}&
		\Take_i^{\Assign_i = 0} = 0 \,\wedge\, \Take_i^{\Assign_i = 1} = 0 \,.
	\end{eqnarray*}
	
Now we can define the individual treatment effect (ITE) for each individual $i$ and the local average treatment effect (LATE) for the compliers:
	\begin{eqnarray*}
	\textrm{ITE}_i 
	&=_\text{df}& \Cure_i^{\Take_i = 1} - \Cure_i^{\Take_i = 0} \,.
	\\[1em]
	\textrm{LATE} 
	&=_\text{df}& \frac{\sum_{i: \, \Complier(i)} \textrm{ITE}_i}{ \#\{ i :  \Complier(i) \} }
	\,.
	\end{eqnarray*}	
To make the LATE well-defined, the denominator must be assumed to be nonzero:
	\opp  		
	{\bf Assumption 5 (Existence of Compliers)} $\Complier(i)$ for some individual $i$. 
	\edd  

The design of the cards itself is non-probabilistic. In the Rubin causal model, probabilities arise entirely from how individuals are drawn from the population and assigned to different groups. For simplicity, let the subscript-free notation $\PP\left( \Cure^{\Take = 0} = 1 \right)$ denote the probability of drawing an individual from the population who would be cured without taking the treatment. If everyone has an equal probability $1/N$ of being selected, where $N$ is the population size, then $\PP\left( \Cure^{\Take = 0} = 1 \right)$ is identical to the proportion of those who would be cured without taking the treatment. This exploits a convenient ambiguity of $\PP$ between {\em pr}obability and {\em pr}oportion. There are two probabilistic assumptions:
	\opp  		
	{\bf Assumption 6 (Random Selection).} Everyone in the population has an equal probability of being selected. In other words, the probability
	$$\PP
	\left(
	\Take^{\Assign = 0} = a, \,
	\Take^{\Assign = 1} = b, \,
	\Cure^{\Take = 0} = c, \,
	\Cure^{\Take = 1} = d \right) $$
	is equal to the proportions of the individuals with the corresponding counterfactual properties $\Take^{\Assign = 0} = a$, $\Take^{\Assign = 1} = b$, $\Cure^{\Take = 0} = c$, and $\Cure^{\Take = 1} = d$.
		
	{\bf Assumption 7 (Random Assignment).} Any individual, once selected, has a nontrivial probability (say 50\%) of being assigned to the treatment/control group, independently of their identity. So, $\Assign$ is probabilistically independent of the set of all the four potential outcomes in use,
	$\Take^{\Assign = 0}$,
	$\Take^{\Assign = 1}$,
	$\Cure^{\Take = 0}$, and
	$\Cure^{\Take = 1}$; in symbols:
	\begin{align*}
	& \PP
	\left(
	\Take^{\Assign = 0} = a, \,
	\Take^{\Assign = 1} = b, \,
	\Cure^{\Take = 0} = c, \,
	\Cure^{\Take = 1} = d \right)
	\\[0.5em]
	& =\; \PP
	\left(
	\Take^{\Assign = 0} = a, \,
	\Take^{\Assign = 1} = b, \,
	\Cure^{\Take = 0} = c, \,
	\Cure^{\Take = 1} = d \given Assign = 0 \right)
	\\[0.5em]
	& =\; \PP
	\left(
	\Take^{\Assign = 0} = a, \,
	\Take^{\Assign = 1} = b, \,
	\Cure^{\Take = 0} = c, \,
	\Cure^{\Take = 1} = d \given Assign = 1 \right) \,.
	\end{align*}
	\edd  
There is one final assumption:
	\opp  		
	{\bf Assumption 8 (No Defiers)} $\Defier(i)$ for no individual $i$. 
	\edd  
This assumption is presented last because, in real applications, it is often the one most responsible for delineating the scope of the method of instrumental variable estimation. 

Then we have the classic result due to Imbens \& Angrist (1994) and Angrist, Imbens, \& Rubin (1996):
	\opp 
	{\bf Theorem 1 (Formal Version).} Under the assumptions 1-8 as stated above,
	\begin{eqnarray*}
	{\rm LATE}
	&=& \frac{
	\PP\left(\Cure = 1\given \Assign = 1\right) \,-\, \PP\left(\Cure = 1 \given \Assign = 0\right)
	}{
	\PP\left(\Take = 1 \given \Assign = 1\right) \,-\, \PP\left(\Take = 1 \given \Assign = 0\right)
	} \,.
	\end{eqnarray*}
	\edd 
I believe that this list of assumptions, 1-8, is the most comprehensive one currently available.

\subsection{Proof of the Main Result: Theorem 2}\label{app-proof-2}

Recall that each individual $i$ has an individual treatment effect given by ${\rm ITE}_i = c_i - c^*_i$, with a degree of compliance defined by ${\rm DC}_i = t_i - t^*_i$. Hence the DATE can be expressed as follows:
	\begin{eqnarray*}
	{\rm DATE} 
	&=& \sum_{i: \text{ being a complier}} 
	\;
		\underbrace{
			\left( 
			\frac{{\rm DC}_i}{\sum_{j: \text{ being a complier}} {\rm DC}_j} 
			\right)
		}_{
			\mbox{$=$ the weight of $i$} 
		}
	\;
	{\rm ITE}_i
\\
	&=& \sum_{i} 
	\;
			\left( 
			\frac{{\rm DC}_i}{\sum_{j} {\rm DC}_j} 
			\right)
	\;
	{\rm ITE}_i
	\quad = \quad 
	\sum_{i} 
			\left( 
			\frac{t_i - t^*_i}{\sum_{j} \big(t_j - t^*_j\big)} 
			\right)
	\big( c_i - c^*_i \big) \,.
	\end{eqnarray*}
The first line is just the definition of the DATE, which is well-defined (with a nonzero denominator) by the assumption of Existence of Compliers*. In the second line, $i$ and $j$ are no longer restricted to compliers but range over all individuals in the population; this is justified by the assumption of No Defiers* and by the fact that indifference-takers carry zero weights. Now, the goal is to verify this equation:
	\begin{eqnarray*}
	{\rm DATE}
	&\overset{?}{=}&
	\frac{
		\PP\left( \Cure = 1 \given \Assign = 1  \right)
		\,-\, \PP\left( \Cure = 1 \given \Assign = 0  \right)
	}{
		\PP\left( \Take = 1 \given \Assign = 1  \right)
		\,-\, \PP\left( \Take = 1 \given \Assign = 0  \right)
	} \,.
	\end{eqnarray*}
The terms on the right-hand side are to be calculated in turn. I will leverage a defining feature of the causal Bayes net, the {\em Causal Markov Assumption}, which asserts that every variable is probabilistically independent of its non-descendants (non-effects) given its parents (direct causes). Start with the first term in the numerator. By applying the Chain Rule, we have:
	\begin{eqnarray*}
	\PP\left( \Cure = 1 \given \Assign = 1  \right)
	&=& \sum_{i,j} \Big(
		\PP \,( 
		\Cure = 1 \given 
		\Take = j, \Individual = i, 
		\Assign = 1
		) 
	\\[-0.5em]
	&& \quad\quad\quad\quad\cdot\;  
		\PP\left( 
		\Take = j \given 
		\Individual = i, \Assign = 1
		\right)
	\\
	&& \quad\quad\quad\quad\cdot\;  
		\PP\,(  
		\Individual = i \given 
		\Assign = 1
		)
	\Big)
	\end{eqnarray*}
The above can be simplified by the Causal Markov Assumption:
	\begin{eqnarray*}
	\PP\left( \Cure = 1 \given \Assign = 1  \right)
	&=&
		\sum_{i,j} \Big(
		\PP \,( 
		\Cure = 1 \given 
		\Take = j, \Individual = i, 
			\cancel{\Assign = 1}
		) 
	\\[-0.5em]
	&& \quad\quad\quad\quad\cdot\;  
		\PP\left( 
		\Take = j \given 
		\Individual = i, \Assign = 1
		\right)
	\\
	&& \quad\quad\quad\quad\cdot\;  
		\PP\,(  
		\Individual = i \given 
			\cancel{\Assign = 1}
		)
	\Big)
	\end{eqnarray*}
Then, by plugging in the parameters, we have:
	\begin{eqnarray*}
	\PP\left( \Cure = 1 \given \Assign = 1  \right)
	&=&
	\sum_{i} \Big( c_i \, t_i \, \tfrac{1}{N} \Big) \;+\;  \sum_{i} \Big(c^*_i \, (1 - t_i) \, \tfrac{1}{N} \Big)
	\\
	&=& \tfrac{1}{N} \sum_{i} \Big( c_i \, t_i  + c^*_i \, (1 - t_i) \Big) \,.
	\end{eqnarray*}
Similarly for the second term in the numerator:
	\begin{eqnarray*}
	\PP\left( \Cure = 1 \given \Assign = 0  \right)
	&=&
	\tfrac{1}{N} \sum_{i} \Big( c_i \, t^*_i  + c^*_i \, (1 - t^*_i) \Big) \,.
	\end{eqnarray*}
Now calculate the first term in the denominator:
	\begin{eqnarray*}
	\PP\left( \Take = 1 \given \Assign = 1  \right)
	&=&	
		\sum_{i} \;
		\PP\left( 
		\Take = 1 \given 
		\Individual = i, \Assign = 1  
		\right) 	
	\cdot   
		\PP\,( 
		\Individual = i 
		\given 
		\underset{
			\rm by \, Causal \, Markov
			}{
			\cancel{\Assign = 1}
			}  
		)
	\\
	&=& \sum_{i} \,t_i \, \tfrac{1}{N}
\\
	&=& \tfrac{1}{N} \sum_{i} \,t_i \,.
	\end{eqnarray*}
Similarly for the second term in the denominator:
	\begin{eqnarray*}
	\PP\left( \Take = 1 \given \Assign = 0  \right)
	&=& \tfrac{1}{N} \sum_{i} \,t^*_i \,.
	\end{eqnarray*}
To finish off, plug the four terms just calculated into the following:
	\begin{align*}
	&\frac{
		\PP\left( \Cure = 1 \given \Assign = 1  \right)
		\,-\, \PP\left( \Cure = 1 \given \Assign = 0  \right)
	}{
		\PP\left( \Take = 1 \given \Assign = 1  \right)
		\,-\, \PP\left( \Take = 1 \given \Assign = 0  \right)
	}
\\[0.5em]
	&=\;
	\frac{
		 \cancel{\tfrac{1}{N}} \sum_{i} \Big( c_i \, t_i  + c^*_i \, (1 - t_i) \Big)
		 -
		  \cancel{\tfrac{1}{N}} \sum_{i} \Big( c_i \, t^*_i  + c^*_i \, (1 - t^*_i) \Big)
	}{
		 \cancel{\tfrac{1}{N}} \sum_{i} \,t_i
		 -
		 \cancel{\tfrac{1}{N}} \sum_{i} \,t^*_i
	}
\\
	&=\;
	\frac{
		 \sum_{i} \Big( c_i\,t_i  + \cancel{c^*_i} - c^*_i\,t_i - c_i\,t^*_i - \cancel{c^*_i} + c^*_i\,t^*_i  \Big)
	}{
		\sum_{i} \,t_i
		 -
		\sum_{i} \,t^*_i
	}
\\
	&=\;
	\frac{
		 \sum_{i} \big( t_i - t^*_i \big)\big( c_i - c^*_i \big)
	}{
		 \sum_{j} \big(t_j - t^*_j\big)
	}
\\
	&=\; 
	\sum_{i} 
			\left( 
			\frac{t_i - t^*_i}{\sum_{j} \big(t_j - t^*_j\big)} 
			\right)
	\big( c_i - c^*_i \big)
	\;\;=\;\;
	{\rm DATE} \,.
	\end{align*}
Q.E.D.


\section*{References}
\begin{description}	
	\item Angrist, J. D. (1990) ``Lifetime Earnings and the Vietnam Era Draft Lottery: Evidence from Social Security Administrative Records'', {\em American Economic Review}, 80, 313-336.
	
	\item Boylan, D. (2024) ``Counterfactual Skepticism Is (Just) Skepticism'', {\em Philosophy and Phenomenological Research}, 108(1), 259-286.
	
	\item Christensen, D. (1997) ``What Is Relative Confirmation?'', {\em No\^{u}s}, 31(3), 370-384.

	\item Dawid, A. P. (2000) ``Causal Inference without Counterfactuals'', {\em Journal of the American Statistical Association}, 95(450): 407-424.
	
	\item Emery, N. (2017) ``The Metaphysical Consequences of Counterfactual Skepticism'', {\em Philosophy and Phenomenological Research}, 94(2), 399-432.
	
	\item Field, H. (2016) {\em Science without Numbers}, Oxford University Press.
	
	\item H\'{a}jek, A. (unpublished manuscript) ``Most Counterfactuals Are False'', URL = \\$<$https://philarchive.org/rec/HJEMCA$>$
	
	\item Hausman, D. M. (2024) ``Philosophy of Economics'', Zalta, E. N. \& Nodelman, U. (eds.) {\em The Stanford Encyclopedia of Philosophy} (Fall 2024 Edition), URL = \\$<$https://plato.stanford.edu/archives/fall2024/entries/economics/$>$

	\item Hern\'{a}n, M. A. \& Robins, J. M. (2023) {\em Causal Inference: What If}, Chapman \& Hall/CRC.
	
	\item Hitchcock, C. (2024) ``Causal Models'', Zalta, E. N. \& Nodelman, U. (eds.) {\em The Stanford Encyclopedia of Philosophy} (Summer 2024 Edition), URL = \\$<$https://plato.stanford.edu/archives/sum2024/entries/causal-models/$>$

	\item Imbens, G. W., \& Angrist, J. (1994) ``Identification and Estimation of Local Average Treatment Effects'', {\em Econometrica} 62, 467-476.
	
	\item Imbens, G. W., \& Rubin, D. (2015) {\em Causal Inference for Statistics, Social, and Biomedical Sciences}, Cambridge University Press.

	\item Lewis, D. K. (1973) {\em Counterfactuals}, Blackwell.
	
	\item Longino, H. E. (1979) ``Evidence and Hypothesis: An Analysis of Evidential Relations'', {\em Philosophy of Science}, 46(1), 35-56.

	\item Mandelkern, M. (2022) ``Modals and Conditionals'', in Altshuler, D. (ed.) {\em Linguistics Meets Philosophy}, Oxford University Press, pp. 502-533.
	
	\item Markus, K. A. (2021) ``Causal Effects and Counterfactual Conditionals: Contrasting Rubin, Lewis and Pearl'', {\em Economics \& Philosophy}, 37(3), 441-461.


	\im Palacios, P. (2024) ``Intertheory Relations in Physics'', Zalta, E. N. \& Nodelman, U. (eds.) {\em The Stanford Encyclopedia of Philosophy} (Spring 2024 Edition), URL = \\$<$https://plato.stanford.edu/archives/spr2024/entries/physics-interrelate/$>$
	
	\item Pearl, J. (2000) ``Comment on Dawid's Causal Inference without Counterfactuals'', {\em Journal of the American Statistical Association}, 95(450): 428-431.
	
	\item Pearl, J. (2009), {\em Causality}, Cambridge University Press.

	\item Putnam, H. (1968) ``Is Logic Empirical?'', in Cohen, R. S. \& Wartofsky, M. W. (eds.) {\em Boston Studies in the Philosophy of Science}, Vol. 5, D. Reidel: 216-241. 
	
	\item --------- (1971) {\em Philosophy of Logic}, Routledge.	
	
	\item Quine, W. V. (1948) ``On What There Is'', {\em Review of Metaphysics}, 2(5): 21-38.
	
	\item --------- (1951) ``Two Dogmas of Empiricism'', {\em Philosophical Review}, 60: 20-43.
	
	\item Reiss, J. (2020) ``What Are the Drivers of Induction? Towards a Material Theory$^+$'', {\em Studies in History and Philosophy of Science Part A}, 83, 8-16.
	
	\item Rubin, D. B. (1974) ``Estimating Causal Effects of Treatments in Randomized and Nonrandomized Studies'', {\em Journal of Educational Psychology} 66: 688-701.
	
	\item Spirtes, P., Glymour, C. N., \& Scheines, R. (2000) {\em Causation, Prediction, and Search}, MIT Press.
	
	\item Stalnaker, R. C. (1968) ``A Theory of Conditionals'', in Harper, W. L., Pearce, G. A., \& Stalnaker, R. C. (eds.) {\em Ifs: Conditionals, Belief, Decision, Chance and Time}, Springer Netherlands: 41-55.
	
	\item Stalnaker, R. (1981) ``A Defense of Conditional Excluded Middle'', in: Harper, W. L., Pearce, G. A., \& Stalnaker, R. (ends), {\em Ifs: Conditionals, Belief, Decision, Chance and Time}, D. Reidel Publishing Company, pp. 87-104

	\item Weinberger, N. (2023) ``Comparing Rubin and Pearl's Causal Modelling Frameworks: A Commentary on Markus (2021)'', {\em Economics \& Philosophy}, 39(3), 485-493.
	
	\item Williams, J. R. G. (2010) {\em Defending Conditional Excluded Middle}, No\^{u}s, 44(4), 650-668.
\end{description}

\end{document}